\pdfoutput=1

\documentclass[11pt]{article}

\usepackage[final]{acl}
\usepackage{graphicx}

\usepackage{times}
\usepackage{latexsym}

\usepackage[T1]{fontenc}

\usepackage[utf8]{inputenc}

\usepackage{microtype}

\usepackage{algorithm}
\usepackage{amsmath}
\usepackage{algorithmic}

\usepackage{amssymb} 

\title{Using RL to Identify Divisive Perspectives Improves LLMs Abilities to Identify Communities on Social Media}

\author{Nikhil Mehta \\ Department of Computer Science \\ Purdue University \\ West Lafayette, IN 47907 \\ {\tt mehta52@purdue.edu} \\\And Dan Goldwasser \\ Department of Computer Science \\ Purdue University \\ West Lafayette, IN 47907 \\ {\tt dgoldwas@purdue.edu} \\}

\begin{document}
\maketitle
\begin{abstract}
The large scale usage of social media, combined with its significant impact, has made it increasingly important to understand it. In particular, identifying user communities, can be helpful for many downstream tasks. However, particularly when models are trained on past data and tested on future, doing this is difficult. 

In this paper, we hypothesize to take advantage of Large Language Models (LLMs), to better identify user communities. Due to the fact that many LLMs, such as ChatGPT, are fixed and must be treated as black-boxes, we propose an approach to better prompt them, by training a smaller LLM to do this. We devise strategies to train this smaller model, showing how it can improve the larger LLMs ability to detect communities. Experimental results show improvements on Reddit and Twitter data, on the tasks of community detection, bot detection, and news media profiling.
\end{abstract}

\section{Introduction}
\label{sec:intro}
The rise of social media platforms over the last decade has had a tremendous impact on people's lives, affecting their perspectives on key events such as political elections~\cite{mitchell2016modern, shu2019beyond} and led to the creation of segregated information communities, also known as \textit{``echo chambers''}~\cite{gentzkow2011ideological,quattrociocchi2016echo,dubois2018echo,garimella2018political}.  Following the 
the principal of \textit{social homophily}~\cite{mcpherson2001birds,bessi2016homophily}, 
these tightly-knit communities consist of like-minded users, which have similar viewpoints and content preferences. 

Identifying these information communities can lead to better performance in a number of important social media related downstream tasks, such as news media profiling (fake news and political bias detection), user content recommendation, trend prediction, crisis monitoring, sentiment analysis, and more \cite{bedi2016community}. For example, for media profiling, groups of users sharing left-biased news in the past, are likely to do so in the future.

The community identification task is typically formulated as a form of graph analysis, either predicting missing edges (i.e., friendship relationships), graph clustering (i.e., community detection), or more recently with deep learning, such as using graph neural networks (GNN) \cite{liu2020deep}. However, due to the diversity of content found on social media, understanding users' perspectives using a fixed training set is highly challenging. For example, in the settings of \textbf{emerging news events}, the system is evaluated on its ability to adapt to new events, consisting of previously unseen users and topics. This temporal and topic shift at test time, hurts the performance of many models, and they must be retrained \cite{zhang2023vibe}. Since these settings are highly realistic (new topics and events emerge on social media everyday), we focus this paper on them and we evaluate these settings across a range of social media-related tasks.

\begin{figure}[t!]
  \centering
  \includegraphics[scale=0.4]{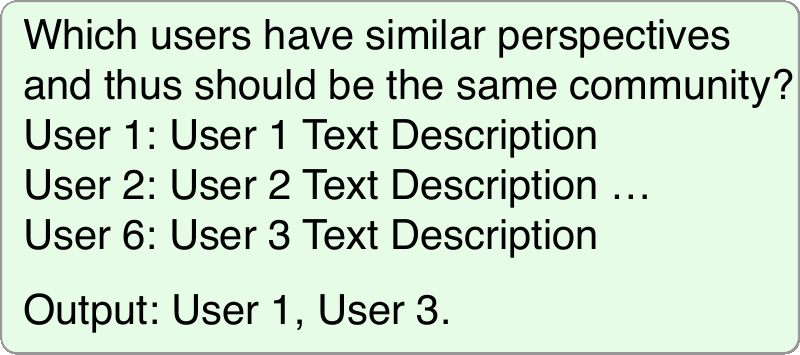}
  \caption{\small An example of the LLM Community Detection Task: Given a set of users and their textual descriptions, determine which users are similar and have similar perspective.}
  \label{fig:chatgpt_comm}
  \vspace{-15pt}
\end{figure}

In this paper, we explore a new direction for tackling such social inference tasks, inspired by the recently popular Large Language Models (LLMs), such as ChatGPT \cite{openai2022}, which perform well on many NLP tasks. Specifically, given their ability to assess textual similarity well \cite{OpenAI2023GPT4TR, li2023deelm}, we ask --  \textbf{\textit{can the strong textual similarity performance extend to the task of community detection?}} Given a set of users and text describing their viewpoints, we explore whether LLMs can identify if any of the users are similar. This way, social inference is reduced to a simpler text similarity problem (comparing user's text descriptions), and LLMs can help us form information communities. Fig.~\ref{fig:chatgpt_comm} shows an example of this community detection LLM task.

Intuitively, given their massive training datasets, LLMs have the potential to generalize across time periods and events, identify users with similar viewpoints, and thus perform well in the important emerging news events settings.  However, we find that this task is still difficult for LLMs. We noticed that LLMs often focus on the high-level aspects of users to determine if they belong in the same community, favoring similarity of interest topics rather than nuanced opinions about them. As a result, LLMs often do not form meaningful communities. For example, two users discussing a popular entity like ``Donald Trump'', could be considered similar by a LLM, when in reality it's the context and attitudes expressed towards ``Donald Trump'' that makes them similar or not. If instead the LLM focused on how the users discuss Donald Trump (for example, their opinions on Trump's perspective on issues like gun control) then the LLM could correctly separate users into meaningful communities.

\begin{figure}[t!]
  \centering
  \includegraphics[scale=0.35]{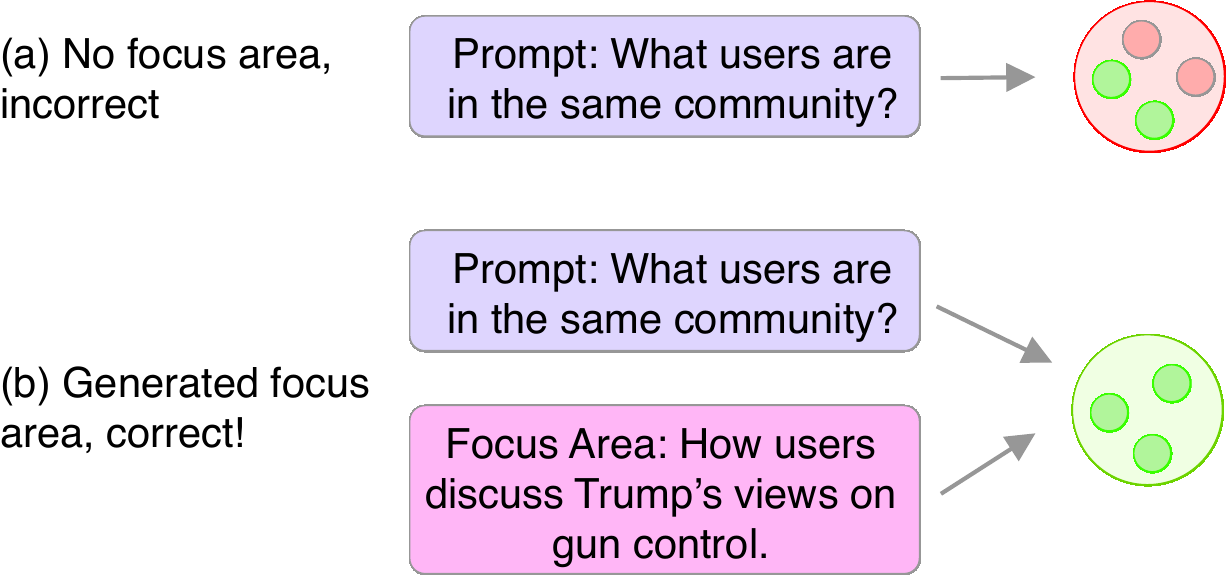}
  \caption{\small An example of how Focus Areas can help. Without them (a), the LLM incorrectly forms the community (red users), but with them (b), the LLM focuses on the divisive issues and correctly forms the community (green).}
  \label{fig:framework_overview}
  \vspace{-15pt}
\end{figure}

Our key technical contribution follows this intuition. We hypothesize that \textbf{focusing the LLM on the relevant aspects of users would result in better information communities}. We propose several models for automatically adding to the LLM  prompt the exact topics and entities it should focus on to separate users into an information community. With the help of this additional information, the LLM can compare the user descriptions, focus on the divisive issues, and form the correct community. We call this additional prompt sentence a \textbf{focus area}. For example, in the running Donald Trump example, the focus area could be: \textit{Focus on how the users discuss Donald Trump's views on gun control}. Tab.~\ref{tab:focus_area_examples} show more ex. of good focus areas, and Fig.~\ref{fig:framework_overview} shows how they can be useful.

Since many of the best performing LLMs are only accessible through an API, or are too large for task-specific training, we treat these models as black-boxes, and train a smaller LM to generate the focus area. This approach offers several advantages, such as being directly usable on top of any LLM, without changing the LLMs performance. We compare several variants of our approach, using the LLM directly (without focus areas), using the LLM to generate the focus areas and finally, training the smaller LM to generate the focus areas and augment the LLM prompts. We train the smaller LM using Reinforcement Learning (RL). The reward signal used by the RL algorithm is obtained by combining several rewards, such as the performance of the LLM when using the generated focus areas and ``unsupervised'' metrics capturing focus area topic relevance, informativeness, impact, length, and more (see Sec.~\ref{subsubsec:reward_functions} for details).

We evaluate our approach in two settings. First, we define an intrinsic evaluation over Reddit and Twitter data, where users are sampled from known communities. Our goal is to recover the ground-truth community memberships via the focus-area augmented LLM prompts. Second, we look at the contribution of a focus-area augmented LLM based approach for downstream tasks that require social information -- identifying false information and political bias in news media. Here, the gold community membership is unknown and can only be gauged by its contribution the downstream task. In both settings we model the out-of-domain emerging news event settings, by training the focus-area generator on a single community, and using it to generate focus areas for new, unseen communities.

In short, we make the following contributions: \textbf{(1)} We propose to use large, frozen LLMs to detect information communities on social media. \textbf{(2)} We train a smaller LM to generate a focus area, an additional prompt sentence to feed into the bigger LLM, to better detect information communities. To train the LM, we devise a novel Supervised and RL training procedure, specific to the social media setting. \textbf{(3)} We show how better community detection can improve the performance of downstream social media tasks in the challenging settings of emerging news events, specifically community detection, bot detection, and news source profiling (factuality/bias detection). We use Reddit and Twitter data.

Sec.~\ref{sec:model} describes our framework, Sec.~\ref{sec:experiments} our results, Sec.~\ref{sec:discussion} analyzes, and Sec.~\ref{sec:summary} concludes.

 \section{Related Work}
 \label{sec:related_work}
 Over the last few years, there has been a number of works analyzing social media, whether it is news media profiling \cite{baly:2018:EMNLP2018, baly:2020:ACL2020}, fake news detection \cite{mehta2022tackling, yang2023entity}, Reddit analysis \cite{arazzi2023importance}, Bot Detection \cite{tan2023botpercent}, or topic analysis \cite{roy2023tale}. These works utilize a variety of ML frameworks, such as LLMs \cite{su2023adapting} and graphs \cite{phan2023fake, ali2023social}, and evaluate a variety of settings such as cross-domain \cite{shu2022cross} and low-resource \cite{lin2022detect} ones. A more realistic and more challenging setting to analyze, which we also do, is one in which test samples mention different topics and feature different users than seen at training time. Due to their importance, these settings have also recently received more attention \cite{zhang2023vibe, mehta2023interactive, mehta2023interactively}.

An important part of social media analysis is analyzing the users on social media. Specifically, prior work \cite{bessi2016homophily, ali2023social} shows how grouping the users into information communities can provide insight for downstream tasks, such as fake news detection \cite{mehta2022tackling}, content recommendation \cite{singh2022social}, or even general analysis 
\cite{aguilar2022social} such as how users view major events\cite{hao2024social}. In general, understanding user perspectives and forming these communities, is important, see: App.~\ref{app:importance_community_detection}. 

Large Language Models (LLMs) have been applied to a large amount of social media related tasks, like fake news detection \cite{su2023adapting}, as they can capture a large amount of knowledge learned from their extensive pre-training. While they can succeed at many NLP tasks like summarization \cite{pu2023summarization}, they still struggle on reasoning tasks like needed for social media analysis. However, as we later show, when appropriately prompted, their performance on these tasks improves.

Prompting LLMs has been studied in a variety of ways, whether it be chain-of-thought reasoning \cite{wei2022chain}, chain-of-hindsight \cite{liu2023chain}, self-refinement \cite{madaan2023self} or RLHF \cite{sun2023reinforcement}. Similar to \citeauthor{akyurek2023rl4f}, we aim to train a smaller language model to prompt bigger, frozen LLMs. Similarly, improving LLMs using feedback has also received increasing research attention, across a variety of tasks, such as summarization \cite{ma2023eureka, yao2023improving, hu2023aligning}. However, compared to tasks like summarization and action planning, social media analysis requires a more nuanced analysis, which affects the way we train our models (i.e. reward functions), and the feedback we provide.

 \section{Model}
 \label{sec:model}
 Our goal in this paper is to improve big, frozen LLMs performance on social media related tasks, specifically detecting user communities, as described in Sec.~\ref{subsec:community_detection_definition}. To do this, we train a smaller LLM to add additional text, which we call a ``focus area'' (Sec~\ref{subsec:focus_area_definition}), to the prompt of the bigger one. We train the smaller model first using Supervised Learning (Sec.~\ref{subsec:supervised_training}), and then Reinforcement Learning (Sec.~\ref{subsec:reinforcement_learning}). Similar to \citeauthor{akyurek2023rl4f}, we refer to the bigger, frozen LLM as $\text{LLM}_{\text{task}}$, and the smaller one as $\text{LLM}_{\text{prompt}}$.

\subsection{User Community Detection}
\label{subsec:community_detection_definition}
As mentioned in Sec.~\ref{sec:intro}, detecting user communities has many advantages, such as understanding social media, content recommendation, etc. Moreover, using frozen LLMs to do this can bring further benefits, such as generalizing to new domains, avoiding fine-tuning big models, etc. Thus, in this section, we describe how we formulate the community detection task for frozen LLMs.

As the big, frozen, $\text{LLM}_{\text{task}}$ model can't be trained, it must be prompted. However, LLMs have limited context size, so we cannot prompt them with all the users on social media. Thus, we instead define the following, more simplified \textbf{community detection task}, which can be extended: Given a set of six users $U = {u_1, ... u_6}$, each with a textual description describing them, determine which, if any, users are similar to each other and should be in the same community $c_1 = {u_1, ... u_c}$.

$\text{LLM}_{\text{task}}$ responds in natural language, listing the users that are in the same community, and the ones that aren't. Fig.~\ref{fig:chatgpt_comm} shows a shortened example of this task, including our prompt to $\text{LLM}_{\text{task}}$, and App.~\ref{appendix:community_detection} provides details (including generalization). 

The textual description of each user in the prompt to $\text{LLM}_{\text{task}}$ is formed based on their social media posts, and provides information to $\text{LLM}_{\text{task}}$ to help it determine the user similarity. To form it, we prompt Chat-GPT to create a summary of the user given their posts (Twitter tweets, Reddit posts, etc.). We form this summary as it simplifies the community detection process, capturing the key details of the users viewpoints, while also being simpler to analyze than the individual posts. To ensure a relevant summary, we sample posts from users so that all six users $U$ discuss at least one entity in common. An ex. of the $\text{LLM}_{\text{task}}$ prompt we use is shown below in Tab.~\ref{tab:summarize_users}, Fig.~\ref{fig:chat_gpt_summarize_users}, and App.~\ref{appendix:community_detection}. 

We note that in this setup, we ask $\text{LLM}_{\text{task}}$ to detect a max of one community, placing all other users after, or not in a community. This setup can handle real-world settings, where there may be multiple, one, or no user communities in the users presented to the LLM. If there are multiple, $\text{LLM}_{\text{task}}$ should form the most tightly-knit community.

\begin{table}[h!]
\begin{center}
\begin{tabular}{|c|c|}
  \hline
  {\textbf{\small Format}} & {\textbf{\small Language}}\\
 \hline
  \small Chat-GPT Question & \small \begin{tabular}[x]{@{}c@{}}What is the user discussing \\... what is their perspective?\end{tabular} \\ 
  \hline 
  \small Input & \small Reddit Comment: ... \\
  \hline 
  \small Output & \small The user mentions ...\\
 \hline
\end{tabular}
\caption{\small The question, text, and output format we use to create user summaries using $\text{LLM}_{\text{task}}$ (shown for Reddit). }
\label{tab:summarize_users}
\end{center}

\vspace{-20pt}
\end{table}

\subsection{ \texorpdfstring{$\text{LLM}_{\text{prompt}}$} {LLM prompt} Definition: Focus Areas}
\label{subsec:focus_area_definition}
In order to improve $\text{LLM}_{\text{task}}$'s ability to detect communities, we provide it an additional sentence as part of the prompt, which we call a \textbf{focus area}. The focus area tells the LLM exactly what to focus on when reading the user summaries, in order to properly separate the users into communities. We define this focus area to be a short sentence that details the divisive issues and topics that the current set of users are discussing. The focus area significantly simplifies $\text{LLM}_{\text{task}}$'s job, as it now just has to compare the user summaries based on the issues provided, to determine the community. Moreover, it makes sure $\text{LLM}_{\text{task}}$ does not focus on high-level topics when determining user similarity, but rather on divisive issues. For ex., a focus area could be: \textit{Focus on tax increase in California} (more: Tab.~\ref{tab:focus_area_examples}).

\subsection{\texorpdfstring{$\text{LLM}_{\text{prompt}}$}{LLM prompt}: Supervised Training}
\label{subsec:supervised_training}

To generate the focus areas, we train a smaller LLM, $\text{LLM}_{\text{prompt}}$, similarly to \citeauthor{akyurek2023rl4f}. We initialize it as an encoder-decoder model and fine-tune it to generate focus areas, given user summaries. We use T5-Base \cite{raffel2020exploring}, with 223M params, and then train on gold focus areas.

We approximate the gold focus areas using the gold communities and $\text{LLM}_{\text{task}}$, prompting it to generate the focus area based on the user summaries. Specifically, since we know the gold communities from the training data, we ask $\text{LLM}_{\text{task}}$: \textit{What topics separate the gold communities?} Since $\text{LLM}_{\text{task}}$ is told what the gold communities are, it is able to consider what separates the users to form the gold communities, and generate an initial focus area. We show an example in Tab.~\ref{tab:generate_gold_focus_area}, a detailed example in Fig.~\ref{fig:llm_task_generate_gold_focus_areas}, and provide details in App.~\ref{appendix:gold_focus_area_generation}.

\begin{table}[h!]
\begin{center}
\begin{tabular}{|c|c|}
  \hline
  {\textbf{\small Format}} & {\textbf{\small Language}}\\
 \hline
  \small $\text{LLM}_{\text{task}}$ Question & \small \begin{tabular}{@{}c@{}}
    What topics should we focus on \\
    to determine first 3 users are in a  \\
    community, while others are not?
\end{tabular} \\ 
  \hline 
  \small Input & \small $\text{User}_{1} \text{ Summary, ... User}_{n} \text{ Summary}$ \\
  \hline 
  \small Output & \small Focus on ...\\
 \hline
\end{tabular}
\caption{\small The question, input text, and output format we to create gold focus areas. }
\label{tab:generate_gold_focus_area}
\end{center}

\vspace{-20pt}
\end{table}

\subsection{\texorpdfstring{$\text{LLM}_{\text{prompt}}$}{LLM prompt}: Reinforcement Learning}
\label{subsec:reinforcement_learning}
The supervised training phase above initializes the model to generate focus areas, but unfortunately, due to the gold data, many are still too high-level, and thus can be improved, for better community detection. Further, the gold data used to train $\text{LLM}_{\text{prompt}}$ comes from $\text{LLM}_{\text{task}}$, and our goal is to improve $\text{LLM}_{\text{task}}$'s performance. Thus, we must train $\text{LLM}_{\text{prompt}}$ directly on community detection, which we do using $\text{LLM}_{\text{task}}$'s predicted community outputs, when the output focus area from $\text{LLM}_{\text{prompt}}$ is used. However, as $\text{LLM}_{\text{task}}$ is not trainable, we use Reinforcement Learning (RL), with several novel reward functions (RF), which we design specifically for community detection and describe in Sec~\ref{subsubsec:reward_functions}. We then describe our curriculum learning RL training procedure in Sec.~\ref{subsubsec:curriculum_learning}.

\subsubsection{Reward Functions}
\label{subsubsec:reward_functions}
We use 4 novel reward functions to train $\text{LLM}_{\text{prompt}}$ to generate better focus areas. To optimize them, we use the same training dataset as Sec.~\ref{subsec:supervised_training}.

\textbf{RF1: Coverage, Community Detection Performance:} Our first reward, \textbf{Coverage}, described in detail in Sec.~\ref{subsec:evaluation_llm_task}, optimizes community detection directly, thus learning focus areas that help improve community detection performance. Specifically, given two gold communities $c_1, c_2$, and two predicted communities $p_1, p_2$, the reward is: \textit{How many users from each predicted community are part of the same gold community?} To compute this reward while ignoring the order of predicted communities, we first find the largest overlapping gold community for each predicted one, and then compute the overlap accuracy score for each. We note that while $\text{LLM}_{\text{task}}$ is prompted to predict one community (for simplicity), it still places the rest of the input users together in another community, and we have gold data for two communities, which is why this reward function evaluates both.

\textbf{RF2: Entity Frequency:} Our second reward, entity frequency, improves focus areas by getting them to mention entities that may be useful to separate users into communities. To do this, we find entities that are more frequently mentioned by one gold community compared to the others, and provide a reward based on how many of those entities the focus area mentions. Specifically, we first extract entities (Spacy NER-tagger \cite{Honnibal_spaCy_Industrial}) from each user summary, keeping ones that are mentioned more than once across a gold community. Then, we find the entities that are mentioned more often by one of the gold communities. We provide a reward based on how many of these entities are mentioned in the generated focus area scaled to a max of 3 (i.e. 3+ entities = 1.0 reward).

\begin{figure}
  \centering
  \includegraphics[scale=0.45]{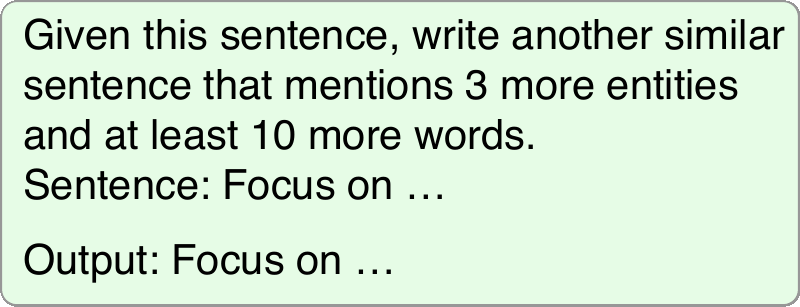}
  \caption{\small How we prompt ChatGPT to generate more informative focus areas (positive class), given ones from the training set (negative class). We then train a binary LR model on this data.}
  \label{fig:chatgpt_complexity}
  \vspace{-15pt}
\end{figure}

\textbf{RF3: Focus Area Informativeness:} This reward function scores focus areas, aiming to make them more informative, so they capture more details about communities. This is essential, as our motivation for providing focus areas to $\text{LLM}_{\text{task}}$ is to make it not rely on general topics, but rather details, to determine communities. To score focus areas, we train a Logistic Regression model on data generated using ChatGPT. We use gold focus areas as negative examples, and for positive examples, we prompt ChatGPT to generate more informative versions of the gold focus areas (as seen in Fig.~\ref{fig:chatgpt_complexity}).

\textbf{RF4: Focus Area Length:} Our final reward function optimizes focus areas to be longer in length, so they can capture more details. We determine the number of words in the predicted focus area, provide 0.5 reward if it is less than 10, 1.0 if it's more than 35, and otherwise a value that scales linearly between 0.5 and 1.0 (up to 35 words).

\subsubsection{Curriculum Learning}
\label{subsubsec:curriculum_learning}
We finetune $\text{LLM}_{\text{prompt}}$ using Proximal Policy Optimization \cite{schulman2017proximal} and the reward functions above, using the implementation by \cite{akyurek2023rl4f, ramamurthy2022reinforcement}.To stabilize the learning of the reward functions from above (Sec.~\ref{subsubsec:reward_functions}), we use curriculum learning. Alg.~\ref{alg:generate_focus_areas} provides pseudo-code for our overall training process, App.~\ref{app:rl_training_details} details of RL + Reward Functions, and App.~\ref{app:curriculum_learning} details of curriculum learning. Our rewards balance each other, i.e. generating useful, entity relevant, informative, and longer focus areas.

\begin{algorithm}
\caption{\textit{Algorithm to Train $\text{LLM}_{\text{prompt}}$ to Generate Focus Areas}}
\begin{algorithmic}[1]
  \small
    \STATE \textbf{Input:} $\text{LLM}_{\text{prompt}}$, $\text{LLM}_{\text{task}}$ (Initialized Prompt Model, Frozen Task Model)
    \STATE \textbf{Input:} Dataset $\sum_{i=1}^{n}  D = {(u_1...u_6, c_1, c_2, f)}$ (Users $u_1,...u_6$ to separate into communities $c_1, c_2$ and Gold Focus Area $f$ to train $\text{LLM}_{\text{prompt}}$ 
    
  \STATE \textbf{Output:} $\text{LLM}_{\text{prompt}}$ (Trained Focus Area Generation Model)

\STATE \textbf{Supervised Training:} Maximize $f: \mathop{{}\mathbb{E}} \left[ \log p_{\theta} (f | u_1, \ldots, u_n) \right]$ (Train $\text{LLM}_{\text{prompt}}$ to generate focus areas) 

\WHILE {not converged}
\STATE Sample mini-batch: $\sum_{i=1}^{n} D = {(u_1...u_6, c_1, c_2)}$
\STATE Generate focus area: $\hat{f} \sim  \text{LLM}_{\text{prompt}} (u_1...u_6)$
\STATE Use Focus area to get community prediction: $\hat{c_1} \sim  \text{LLM}_{\text{task}} (u_1...u_6, \hat{f})$
\STATE Get Reward Based on Community Prediction: $R = \text{Reward}(c_1)$
\STATE Update $\text{LLM}_{\text{prompt}}$ based on reward $R$
\ENDWHILE

\RETURN $\text{LLM}_{\text{prompt}}$ (Trained Focus Area Generation Model)

 \end{algorithmic}
\label{alg:generate_focus_areas}
\vspace{-5pt}
\end{algorithm}
\vspace{-5pt}

 \section{Experiments}
 \label{sec:experiments}
 \subsection{Datasets}
\begin{table}
\begin{center}
\begin{tabular}{|p{2.4cm}|p{0.7cm}|p{0.7cm}|p{0.7cm}|}
  \hline
  {\textbf{\small Dataset}} & {\textbf{\small Train}}
  & {\textbf{\small Val}} & {\textbf{\small Test}}  \\

 \hline
  \small Reddit Politics & \small  2,789 & \small 100 & \small 550  \\
  \small Reddit Economic & \small - & \small - & \small 232  \\
  \small BotPercent & \small - & \small - & \small  155  \\  
  \small Twitter & \small - & \small - & \small 444  \\
  
 \hline
\end{tabular}
\end{center}
\vspace{-10pt}
\captionsetup{justification=centering}
\caption{\small Dataset size statistics. Each sample has 6 users, and all test users are unique across samples.}
\vspace{-15pt}
\label{table:dataset_statistics}
\end{table}

Our goal in this paper is to improve big, frozen LLMs ($\text{LLM}_{\text{task}}$) ability to detect communities. Specifically, given a set of six users with their profile/post summaries, $\text{LLM}_{\text{task}}$ should be able to detect which (if any) users belong to the same community. We now describe our evaluation datasets, including on downstream tasks (\ref{subsubsec:twitter_data}). Tab.~\ref{table:dataset_statistics} shows the number of samples in our different datasets. 

\subsubsection{Reddit}
Our first dataset, collected by us, directly evaluates how well $\text{LLM}_{\text{task}}$ can detect communities. To get the gold data, we use the social media site Reddit.

Reddit is made up of communities called subreddits, each of which consists of posts relating to a central topic, such as ``Politics''. Reddit users make these posts, and other users interact with the posts by commenting or voting on them (up-vote or down-vote). Each subreddit additionally has designated moderators, users who monitor the subreddit, performing actions such as deleting posts that are not relevant to the subreddit. Further, users often down-vote posts that disagree with the ideas of the subreddit. Thus, subreddits and their up-voted content are very similar to real life communities, as they contain similar minded users that discuss topics relevant to the central theme of the subreddit.

Building on this, we hypothesize that users in the same subreddit, who have a positive up-vote score across all their posts in the subreddit, are members of the subreddit's community. Thus, a set of users from one subreddit form one community, and a set of users from another from a different community, and LLMs should be able to tell the difference.

We build two datasets to evaluate this, sampling data from two polarizing subreddits, or communities. The first (Political) dataset is from the ``Democrats'' subreddit and the ``Conservative'' subreddit, while the second (Economic) is from ``Capitalism'' and ``Socialism''. Each dataset sample has six users across two communities (three from the first subreddit/community, and three from the second), which must be separated. To construct each sample, we find two posts, one from each subreddit, that discuss the same topics (made within three weeks of each other and their titles' having at least one entity in common \cite{akbik2019flair}). For each post, we sample three users that belong to the subreddit and comment on the post. As long as their comments have a positive up-vote score, we know that these three users and post is representative of that subreddit's community. After doing this for both subreddits, we obtain a total of six users, three from one subreddit community and three from another, which forms a sample for our dataset. After creating summaries for each user based on their post comments (as discussed in Sec.~\ref{subsec:community_detection_definition}), we can ask the LLM to detect the communities. 

\subsubsection{TwiBot}
\label{subsubsec:twibot_data}
Our second dataset also evaluates how well $\text{LLM}_{\text{task}}$ can detect which users in a given set of six users are in the same community. However, this dataset is from \citeauthor{feng2021twibot}, and evaluates whether Twitter users are bots or not. The dataset, named TwiBot-20, consists of Twitter users, their metadata (tweets, profile information), and a label signifying whether they are bots or not. The dataset additionally groups users into four broad categories: Politics, Business, Entertainment, and Sports. We construct test samples, each with six users from two communities, using this dataset, where each sample has users belonging to the same category, and the two communities are bot and not bot. While other works \cite{feng2022twibot, tan2023botpercent} also used this dataset, we do not compare to them directly, as our setup is unique to our task (other works use graphs, etc. which we evaluate in Sec.~\ref{subsubsec:twitter_data}). 

\begin{table*}
\begin{center}
\begin{tabular}{|p{8.5cm}|p{1.4cm}|p{2.2cm}|}
  \hline
  {\textbf{\small Dataset: Model}} & {\textbf{\small Coverage}} & {\textbf{\small \# Test Samples}} \\

 \hline
 \small Reddit Political: No Focus Areas  & \small 42.01  & \small 550 \\
 \small Reddit Political: Gold (ChatGPT) Focus Areas & \small 44.66   & \small 550 \\
 \small Reddit Political: $\text{LLM}_{\text{prompt}}$ Focus Areas: Supervised Learning & \small 45.48  & \small 550 \\
 \small Reddit Political: $\text{LLM}_{\text{prompt}}$ Focus Areas: RL Curriculum Learning & \small \textbf{47.85}  & \small 550 \\
 \hline 
 \small Reddit Economic: No Focus Areas & \small  42.25 & \small 232 \\
 \small Reddit Economic: Gold (ChatGPT) Focus Areas & \small 44.60  & \small 232 \\
 \small Reddit Economic: $\text{LLM}_{\text{prompt}}$ Focus Areas: Supervised Learning  & \small 44.00  & \small 232 \\
 \small Reddit Economic: $\text{LLM}_{\text{prompt}}$ Focus Areas: RL Curriculum Learning & \small \textbf{45.58} & \small 232 \\
 \hline 
 \small TwiBot: No Focus Areas & \small 21.63 & \small  155 \\
 \small TwiBot: Gold (ChatGPT) Focus Areas  & \small 19.19 & \small 155\\
 \small TwiBot: $\text{LLM}_{\text{prompt}}$ Focus Areas: Supervised Learning & \small 22.55 & \small 155  \\
 \small TwiBot: $\text{LLM}_{\text{prompt}}$ Focus Areas: RL Curriculum Learning & \small \textbf{22.72} & \small  155 \\
  \hline 

 \hline
\end{tabular}
\end{center}
\vspace{-5pt}
\captionsetup{justification=centering}
\caption{\small Results on Reddit Political, Reddit Economic, and TwiBot (Bot detection \cite{feng2021twibot}) community detection datasets when using ChatGPT for $\text{LLM}_{\text{task}}$ and T5-Base for $\text{LLM}_{\text{prompt}}$. All of this test data is in the unseen emerging news events settings, and features new topics published after the time period the training set was collected from. Using focus areas improves performance on all three datasets, and training  $\text{LLM}_{\text{prompt}}$ using RL leads to the best performance on each dataset. This shows the benefit of our framework to learn useful focus areas, and those focus areas to improve community detection performance, even on domains and time periods not seen at training time.}
\vspace{-10pt}
\label{table:chatgpt_results}
\end{table*}

\subsubsection{News Source Profiling}
\label{subsubsec:twitter_data}
Our final evaluation is on downstream tasks, showing how detecting communities can improve news source profiling (factuality/bias detection). We use the dataset originally proposed by \citet{baly:2020:ACL2020, baly:2018:EMNLP2018} and also evaluated by \citet{mehta2022tackling}.

The dataset consists of sources scraped from Media Bias/Fact Check\footnote{\url{https://mediabiasfactcheck.com}}, each labeled on a 3-point scale for factuality (high, low, mixed) and bias (left, center, right). Following prior work \cite{baly:2020:ACL2020}, we aim to predict the factuality/bias of the news sources using Twitter data, which provides social context. It consists of sources (the classification targets), the articles they publish, and users who interact with the sources or articles (propagate the articles, follow users/sources). Following \citeauthor{mehta2022tackling}, we build an information graph using this data. We follow the challenging fully inductive evaluation protocol proposed by \citet{mehta2023interactive}, where the test set graph is not connected to the training set graph in any way (no users, sources, articles or edges in common). 

Similar to \citeauthor{mehta2022tackling}, we hypothesize that detecting user communities can increase profiling performance. This is because, similar users are likely to have similar views and thus spread similar content, which has similar factuality/bias. This has also been shown in social homophily theory \cite{bessi2016homophily}. Thus, we randomly sample groups of users, ask  $\text{LLM}_{\text{task}}$ to form communities., and connect users in the same communities in the graph.

\subsection{Training/Test Procedure}
We train \textbf{only} on our first Reddit dataset, which consists of politics subreddits: `Democratic' and `Conservative', collected between the start of 2013 and end of 2016. Thus, \textbf{we don't train/finetune on any of the other test datasets}. We provide details in App.~\ref{appendix:training_details}, and release our code and data.

As discussed in Sec.~\ref{sec:intro}, all of our test data is in the challenging \textbf{emerging news events setting}, which consists of topics and time periods not seen at training time. We first test on the two Reddit datasets, which feature posts made between 2018 and the end of 2023, and then TwiBot-20 \cite{feng2021twibot}. Finally, we evaluate news media profiling, which features posts from after 2019. Importantly, this evaluation is also in the fully inductive setting, so the test set graph does not have any users or nodes in common/connected to the training graph. 

\subsection{Evaluation Metrics}
\subsubsection{\texorpdfstring{$\text{LLM}_{\text{task}}$}{LLM task} Evaluation}
\label{subsec:evaluation_llm_task}
To evaluate $\text{LLM}_{\text{task}}$'s ability to detect information communities, we use a comprehensive metric, which we refer to as \textbf{Coverage}. To compute it, we first determine the appropriate gold community. This is important, as $\text{LLM}_{\text{task}}$ is only asked to predict one community, but the gold data has two. To evaluate, we choose the gold community as the one that has the largest number of overlapping users with $\text{LLM}_{\text{task}}$'s predicted community. We then determine how many users were correctly predicted, out of all the users both predicted and missing. Mathematically: 

\begin{equation}
\frac{\text{\# of correct pred.}}{\text{\# of correct} + {\text{incorrect}} + \text{missing pred.}}
\end{equation}

This metric prioritizes both predicting the communities correctly, and not missing any users.

\subsubsection{News Source Profiling}
\label{subsec:evaluation_metrics_profiling}
For source profiling, we evaluate Accuracy and Macro F1 (the dataset is unbalanced) for news sources, using the dataset proposed by \cite{baly:2020:ACL2020} and expanded by \cite{mehta2023interactive} for the inductive test set settings.

\subsection{Results}
\subsubsection{\texorpdfstring{$\text{LLM}_{\text{task}}$}{LLM task} Evaluation}
Tab.~\ref{table:chatgpt_results} shows our results when we use ChatGPT as $\text{LLM}_{\text{task}}$ on the two Reddit datasets (Political and Economic) and TwiBot Bot Detection \cite{feng2021twibot}. Tab.~\ref{table:llama_results} shows results when Llama 2 is used as $\text{LLM}_{\text{task}}$, showing our framework generalizes across LLMs. We evaluate emerging news events, where test data is unseen and collected from time periods after the training data. On each dataset, focus areas lead to significant performance improvements, particularly our $\text{LLM}_{\text{prompt}}$ model after it is trained with RL and Curriculum Learning. 

When evaluated on the same (but future) domain as training, Reddit Political, $\text{LLM}_{\text{prompt}}$'s focus areas lead to a 5.84\% performance improvement in Coverage, with RL providing $\sim$5\% relative improvement. On a different domain, economic data, performance improves 3.33\%, showing the benefit of our framework to transfer to different domains. On Bot Detection, focus areas lead to more than 5\% improvement. Thus, focus areas improve $\text{LLM}_{\text{task}}$ community detection, even on unseen domains.

\subsubsection{News Source Factuality Detection}
\label{subsec:profiling_results}
\begin{table}
\begin{center}
\begin{tabular}{|p{2.9cm}|p{0.55cm}|p{0.55cm}|p{0.55cm}|p{0.55cm}|}
  \hline
  {\textbf{\small Model}} & {\textbf{\small FN Acc.}} & {\textbf{\small FN F1}} & {\textbf{\small Bias Acc.}} & {\textbf{\small Bias F1}} \\

 \hline
 
 \small \cite{mehta2022tackling} & \small 44.66 & \small 28.50 & \small 47.74 & \small 34.69 \\
 \small \cite{mehta2022tackling} + $\text{LLM}_{\text{task}}$ + Focus Areas & \small \textbf{45.53} & \small \textbf{30.17}  & \small \textbf{48.64} & \small \textbf{36.34} \\

 \hline 
\end{tabular}
\end{center}
\captionsetup{justification=centering}
\caption{\small News Source Media Profiling: Fake News (FN) and Political Bias Detection. When added via edges to the graph (1444 edges), the communities formed using ChatGPT for $\text{LLM}_{\text{task}}$ + focus areas lead to  improvements, showing the usefulness of LLMs to form communities which help downstream tasks, despite no training on this domain. }
\vspace{-5pt}
\label{table:news_media_profiling}
\end{table}

Tab.~\ref{table:news_media_profiling} shows results on news source factuality detection. We evaluate 444 sources for factuality (183 high, 131 mixed, 128 low) and bias (202 right, 109 left, 108 center, rest unknown), and 212 comms. We compare to \citeauthor{mehta2022tackling}, but in the emerging news events settings, using the public Black Lives Matter data from \citeauthor{mehta2023interactively}. We see that using $\text{LLM}_{\text{task}}$ with Focus Areas to form communities leads to improvements (over 4\% relative increase on Bias F1). This shows the benefit of using LLMs to form communities to improve downstream social media tasks, particularly when LLMs are prompted with focus areas. Details: App.~\ref{app:profiling_details}.

 \section{Discussion}
 \label{sec:discussion}
 We analyze our best RL $\text{LLM}_{\text{prompt}}$ model, with ChatGPT. We do an ablation study of our reward functions, (~\ref{subsec:ablation_study}), then a human analysis of generated focus areas, (\ref{subsec:human_analysis}), then case studies (App.~\ref{sec:case_study}), then an analysis of LLM detected user communities for factuality detection (App.~\ref{sec:discussion_graph_user_purity}), and finally discuss the real world impact of our approach (~\ref{sec:real_world_impact}).

\begin{table}
\begin{center}
\begin{tabular}{|p{3.3cm}|p{1.4cm}|p{1.4cm}|}
  \hline
  {\textbf{\small Reward Fn.}} & {\textbf{\small Coverage}} & {\textbf{\small \# Samples}} \\

 \hline

 \small None: No Focus Area & \small 42.01 & \small 550 \\ 
 \small Coverage & \small 46.07 & \small 550 \\
 \small Entity Frequency & \small 46.96 & \small 550 \\
 \small Informativeness & \small 46.90 & \small 550 \\
 \small Length & \small 45.58 & \small 550 \\
 \small All: Curriculum Learning & \small \textbf{47.85} & \small 550 \\
 \hline 
\end{tabular}
\end{center}
\captionsetup{justification=centering}
\caption{\small ChatGPT + T5-Base Reward Function Ablation Study on Reddit Political Data. Although each reward function leads to improvements, using all of them via Curriculum Learning performs the best.}
\label{table:ablation_study}
\vspace{-10pt}
\end{table}

\subsection{Ablation Study}
\label{subsec:ablation_study}

App.~\ref{app:rl_impact} shows the benefit of RL, and Tab.~\ref{table:ablation_study} the results of our reward function ablation study. While we notice improvements compared to not using focus areas, they are not as significant, showing the benefit of RL and learning the rewards together. Doing so enables each reward function to contribute to learning an overall useful focus area.

\subsection{Human Analysis of Focus Areas}
\label{subsec:human_analysis}
We have 3 humans analyze 50 of $\text{LLM}_{\text{prompt}}$'s focus areas, comparing them to the ChatGPT generated ones. They score each focus area on a scale of 1-5, for grammatical correctness and usefulness (to identify divisive issues and user communities). On average, on grammar, ChatGPT scores 4.95, and $\text{LLM}_{\text{prompt}}$ 3.00. However, on usefulness, ChatGPT scores  3.07 and $\text{LLM}_{\text{prompt}}$ 3.26. From this, we see $\text{LLM}_{\text{prompt}}$ generates better focus areas to separate users into communities, which explains our results from Sec.~\ref{sec:experiments}. App.~\ref{app:human_analysis} provides details.

\subsection{Real World Impact}
\label{sec:real_world_impact}
Our framework to generate focus areas can be utilized with any $\text{LLM}_{\text{task}}$ in the real world, even without fine-tuning it. This is because, focus areas are just an additional input to the prompt of $\text{LLM}_{\text{task}}$. Moreover, as we evaluated extensively on emerging news events, particularly on topics and tasks on which our models were not trained on (Reddit Economic, TwiBot, and Source Factuality Detection) our framework is very applicable in the real world on social media, where new topics arise daily. Most importantly, $\text{LLM}_{\text{prompt}}$ doesn't have to be retrained every time a new topic arises. 
 
 \section{Conclusion}
 \label{sec:summary}
 In this paper, we proposed to use large, frozen LLMs to detect user information communities on social media, particularly in the challenging settings of emerging news events, where test data features topics and time periods not seen at training time. We then improved this LLMs performance, by training a smaller LM ($\text{LLM}_{\text{prompt}}$) to generate a focus area, an additional sentence to feed into the bigger LLM. This focus area focuses the LLM on the relevant aspects of users that would result in better information communities, such as divisive issues. Experimental results on Reddit and Twitter data showed performance improvements in detecting communities when using Focus Areas, even on emerging news events. Further, we learned meaningful communities, that lead to improvements on the downstream task of source profiling (factuality/bias detection). Our future work is to generate better focus areas, i.e exploring reward functions.

 \section{Ethics Statement}
 \subsection{Limitations}
 \label{sec:limitations}
In this paper, we proposed a framework to train and evaluate on social media data, specifically Reddit and Twitter data and English. The framework we presented, and the experimental results we achieved, are shown for these domains/tasks. We believe that they will generalize to other domains and tasks, but we leave the exploration of that to future work.

In this paper, we focused on the emerging news events settings, where we evaluated when the test data was not seen at test time. These are some of the most challenging settings for social media tasks, as knowledge learned at training time can't always be used at test time. This is also why we leveraged LLMs for this task. Our future work involves testing how our experiments in this paper can generalize to other domains of emerging news events.

In this paper, we used two Large Language Models: ChatGPT and Llama 2. For ChatGPT, we used the API released publicly by OpenAI, and the details of the model are not known. For Llama 2, we ran it locally, using the Llama-cpp-python library. We specifically run the 70B parameter model, as detailed in Appendix~\ref{appendix:training_details}. While we use both of these models as black-boxes, and they perform well in numerous benchmarks \cite{qinChatGPT}, we understand that our frameworks build on these models and this could be a potential limitation. We believe it's important to take caution when deploying these models.

For experimental reasons, we set up our framework to detect communities in sets of 6 users. We hypothesize this can generalize, to number of users more than or less than 6. Specifically, it there are less than 6 users, generalizing is simple, just provide less users in the prompt. If there are more than 6, our framework can be used by either breaking the number of users into groups of 6, and then asking the LLM to detect communities, or by just passing in more than 6 users at once. While we did not test the latter, we hypothesize it may still work provided the LLM has the ability to handle the longer context, and leave it for future work.
 
 \subsection{Ethics}
 \label{sec:ethics}
 We do not believe we violated any code of ethics in our experiments done in this paper. We release our full code and anonymized data, to make the re-implementation of our models as simple as possible. We also caution that our models are the output of a machine learning model, and this could be parameter/machine dependent. 

In our Reddit dataset release, we anonymized all the user data, to violate no code of ethics. Further, the data we scraped was released publicly by \cite{chang2020convokit}. Thus, all the data we used is previously publicly available. 

Our framework in general is to be used to analyze social media and form information communities along with LLMs. Our general experimental settings of forming focus areas may also be useful for other tasks, and we leave the investigation of this to future work.

Our framework also has the potential to be used in malicious ways, along with positive ones. Specifically, identifying users that belong to specific communities can potentially impact those users, even in harmful ways, such as if this knowledge is made public. While there are clear positives to our community detection approach, such as downstream tasks or finding `friends' for other users, this is one of the downsides. Thus, our framework must be used with caution. 

When considering our work, it's important to consider these and other related things to make sure the usage of our framework and code/data release falls within appropriate and safe use.
 
\bibliography{anthology,custom}

\newpage
\appendix

\section{Importance of Community Detection}
\label{app:importance_community_detection}
In this paper, we aimed to improve the performance of LLMs to detect communities of users on social media. Community detection is an important task in social media analysis, for several reasons. For example, if we know from historical data that a group of users have similar perspectives and are thus in the same community, then it's more likely that content shared by some of the users in the group will also be agreed upon by others in the group. This can be beneficial for downstream tasks. For example, for fake news detection, news shared by a community that historically shares fake news is more likely to be fake news. Further, users in that community are also more likely to share fake news. Thus, if we can identify the fake news sharing community, we have more knowledge about the users in that community, and we can identify new fake news content better. Similar ideas apply to political bias detection. This was also shown by \cite{bessi2016homophily, del2016spreading, mehta2022tackling, mehta2023interactive}. 

Prior work has also shown how detecting communities on social media can improve other downstream tasks, beyond fake news/political bias detection. It can help us analyze trends on social media \cite{singh2022social}, such as how people view major events like COVID-19 over time \cite{hao2024social}. It can help us understand how different groups of people are treated, such as female sports fans \cite{fenton2023female}. It can also help us analyze hate speech on social media \cite{ali2023social}, which is important to maintain a healthy society. 

For these reasons and more, the community detection task is very important, which is why we focused on it in this paper. However, as we showed in Sec.~\ref{sec:experiments}, community detection in the out-of-domain settings where test data is never seen before is challenging. As LLMs capture a large amount of external knowledge, and can thus generalize, these out-of-domain settings are where we can take advantage of LLMs to perform better, assuming we use them correctly. This is where focus areas significantly help, as they tell the LLM what topics/entities to focus on, in order to correctly identify communities, unlocking LLMs for this community detection task.

\section{Community Detection Task Details}
\label{appendix:community_detection}
As defined in Sec.~\ref{subsec:community_detection_definition}, our community detection task is: Given a set of six users $U = {u_1, ... u_6}$, each with a textual description describing them, determine which, if any, users are similar to each other and should be in the same community $c_1 = {u_1, ... u_c}$.

\begin{figure*}[t!]
  \centering
  \includegraphics[scale=0.5]{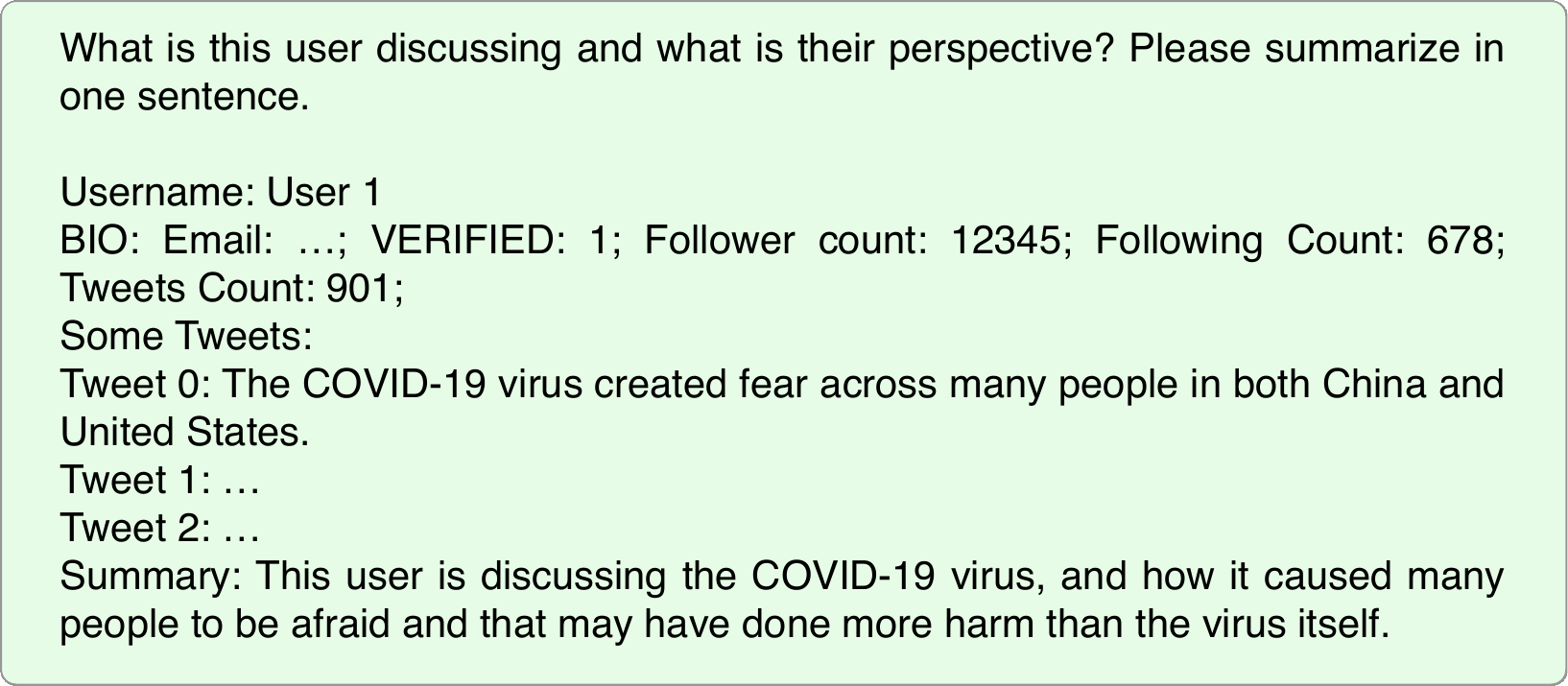}
\caption{\small An example of the prompt we used to determine the user summary. For Twitter users, based on their bio, meta-data, and tweets, we create a summary. For Reddit, we use their comments.}
\label{fig:chat_gpt_summarize_users}
\vspace{-10pt}
\end{figure*}

We represent each user in the prompt to the LLM ($\text{LLM}_{\text{task}}$) by a textual summary, created using Chat-GPT, based on a textual description coming from the user's social media. To get this textual description, for Twitter users, we use 10 of their randomly selected tweets and their profile metadata (profile bio, number of followers, number of people following, number of likes, number of tweets, and whether they are verified or not). For Reddit users, we use the comments they made in relation to the post on the topic $\text{LLM}_{\text{task}}$ is analyzing to determine the community. We provide this textual data to Chat-GPT and ask it to summarize it. The exact summarization task, with prompt input, for Twitter users, is seen in Fig.~\ref{fig:chat_gpt_summarize_users}.

We note that our task setup can easily generalize to any number of users bigger than 6, by breaking them up into groups of 6 and then asking the LLM. More importantly, this task setting is just a way that we set up the input to the LLM. We hypothesize that our entire framework will work with more/less users, and focus areas will be equally effective. Of course, our framework is reliant on $\text{LLM}_{\text{task}}$. Thus, if too many users are used and the LLM cannot handle the large context length, then the baseline and baseline + focus area community detection performance would suffer. We found that 6 was a good number of users for the LLMs we tested.

We also note that it’s possible that there are no user communities in the groups of users presented to $\text{LLM}_{\text{task}}$. In this case, $\text{LLM}_{\text{task}}$ shouldn’t be forced to detect a community. This is why we asked $\text{LLM}_{\text{task}}$ to form only one community, and place all the other users after (i.e. they don’t belong to a community). For example, if there is no community, the LLM won't predict one, and just place all users after (the LLM uses a separator ';;;;;' to separate communities, so in this case the output would be something like: \texttt{$";;;;;user_1, user_2, user_3, etc."$}). On the contrary, if there are multiple communities, $\text{LLM}_{\text{task}}$ should form the single most closely-knit community. This is also a design decision, which future work can change.

Additionally, we note that the community detection process can be done in several steps, i.e. forming a community and narrowing it down in future iterations to be more topically focused, and leave the exploration of this for future work.

\begin{table*}
\begin{center}
\begin{tabular}{|p{8.6cm}|p{1.4cm}|p{1.4cm}|}
  \hline
  {\textbf{\small Dataset: Model}} & {\textbf{\small Coverage}} & {\textbf{\small \# Samples}} \\

 \hline
 \small Reddit Political: No Focus Areas  & \small 58.07  & \small 550 \\
 \small Reddit Political: Gold (Llama 2) Focus Areas & \small 58.20  & \small 550 \\
 \small Reddit Political: $\text{LLM}_{\text{prompt}}$ Focus Areas: Supervised Learning & \small 58.37 & \small 550 \\
 \small Reddit Political: $\text{LLM}_{\text{prompt}}$ Focus Areas: RL Curriculum Learning & \small \textbf{59.21} & \small 550 \\
 \hline
\end{tabular}
\end{center}
\captionsetup{justification=centering}
\caption{\small Results for when we use T5-Base as our $\text{LLM}_{\text{prompt}}$ and Llama 2 as our $\text{LLM}_{\text{task}}$. Results show improvements when using focus areas, a 1.96\% performance increase. Although the improvements are not as significant as when we use ChatGPT as Llama is not a strong enough model to benefit from improved focus areas, we still see a strong improvement.}
\label{table:llama_results}
\end{table*}

\section{News Media Profiling Details}
\label{app:profiling_details}
For news media profiling, where we evaluate news source factuality and bias detection, we use the public information graph model from \cite{mehta2022tackling}, which they originally trained for only news source factuality detection. As we also evaluate political bias source detection, we train the model on both classification objectives. The model uses a Relational GCN to encode a graph structure, training it for the Source Node Classification objective. The Graph consists of three node types: Users, Sources, and News Articles, each with an initial representation (twitter information for users and sources, SBERT embedding article text for articles), which is updated throughout the training process. In the graph, articles are connected to the sources they come from with edges, while users are connected to sources they follow, or other users they follow. Thus, the users provide the social information in the graph, which we aim to better learn, by building better information communities.

Once the graph is trained, we evaluate it in the challenging fully inductive settings, where no test nodes are common with the training set, and no test nodes are connected to training set nodes. We aim to determine if strengthening the user relationships in the graph, i.e. building stronger information communities, can improve the performance on this challenging downstream task. To do this, we sample sets of six users that are close to each other in the learned Graph embedding space (this increases the chance that they discuss similar topics), and run them in our community detection approach. Specifically, we ask $\text{LLM}_{\text{task}}$ to either place some (or none) of these users into a community. For the users $\text{LLM}_{\text{task}}$ thinks are similar, we directly connect them in the information graph, using a user-user edge. We then evaluate our downstream tasks, again.

To build the set of six users, we first randomly sample an initial user. We then find the 20 closest users to this user in the graph embedding space, and randomly sample 5 from them. By sampling from users that are close to this initial user, we increase the likelihood that all six users will discuss similar topics, so focus areas can be more effective. But, we also don't only choose the closest five users every time, to encourage diversity, so the graph can potentially learn user similarity it doesn't already know.

When evaluating this, we see that for both downstream tasks, performance increases when this extra user community information is provided, but only when $\text{LLM}_{\text{task}}$ is used with focus areas and RL (our best model). Without focus areas, the communities formed by $\text{LLM}_{\text{task}}$ are likely incorrect, which is why performance drops, as incorrect user information is being spread in the graph. However, with focus areas, community detection performance improves (as we showed via other experiments where we had ground truth for this), and this leads to direct improvements in the downstream task. This is because, the Graph model leverages the user similarity in the newly formed communities, to determine which sources are likely to be fake news / politically biased, as this user information flows throughout the graph. The results in the main paper (Sec.~\ref{subsec:profiling_results}) show the impact of forming good communities for downstream tasks.

\section{RL Training Details}
\label{app:rl_training_details}
In this section, we provide training details for our RL algorithm, and how exactly we train $\text{LLM}_{\text{prompt}}$ to work with $\text{LLM}_{\text{task}}$. We also discuss the mathematical details of our reward functions.

We initialize $\text{LLM}_{\text{prompt}}$ with an encoder-decoder T5-base, and supervised train it as a text generation task, using the gold focus areas as the training data. Specifically, if $\text{LLM}_{\text{prompt}}$ is parameterized by theta, we maximize $E[\text{log} p_\theta (f | x)]$ where the goal is to generate focus areas $f$.

The second step of our training process is the RL stage, using different reward functions (detailed below). We continue training the policy network from the supervised learning stage ($\text{LLM}_{\text{prompt}}$), but now to maximize the reward from the reward functions, using the KL-regularized Proximal Policy Optimization (PPO) objective \cite{schulman2017proximal}. To do this, we sample batches, get the focus areas from $\text{LLM}_{\text{prompt}}$, pass them along with the user summaries to $\text{LLM}_{\text{task}}$, and then get user communities. We run the communities through our reward functions, compute the reward, and update $\text{LLM}_{\text{task}}$, by maximizing the PPO objective.

We now provide more details of RL4LMs, the public reinforcement learning library we used, as proposed by \citeauthor{ramamurthy2022reinforcement} and used by \citeauthor{akyurek2023rl4f}. We provide an overview, more details can be found in \citeauthor{ramamurthy2022reinforcement}. The RL4LMs library provides an OpenAI gym style API to allow us to easily train our models. In RL4LMs, each environment is viewed as a NLP task, i.e. generating focus areas from user summaries. Thus, there is a dataset $D = {(x, y)}$, where $x$ is a language input (user summaries) and $y$ is a target string (focus areas). Generation is viewed as a Markov Decision Process (MDP), consisting of states, actions, rewards, and transition functions. At each episode in the MDP, i.e. for a given dataset sample, the input x is provided to the model ($\text{LLM}_{\text{prompt}}$), and used as the initial state. An action $a$ is then performed, which in this environment means to generate a token from the vocabulary. The transition function then models this and appends this action to the end of the state. This continues until the episode ends, i.e. when all the tokens are generated. At the end of an episode, a Reward based on the state and the gold focus area is provided. The environment can then be updated using the regularized KL reward, by training via PPO.

To improve the stability of training RL algorithms with NLP methods (i.e. handling large vocabulary sizes), RL4LMs also introduces NLPO: Natural Language PPO. NLPO maintains a masking policy, which is a copy of the current policy, but one that is updated only every $u$ steps. This updating policy provides the original policy with an additional constraint that can help regularize the RL training.

\subsection{Reward Function Details}
\label{app:reward_function_details}

Finally, we provide the mathematical details of each of the reward functions we used, expanding  

\textbf{Coverage:} As outlined in Sec.~\ref{subsec:evaluation_llm_task}, the goal of Coverage is to see how well $\text{LLM}_{\text{task}}$ can detect communities. Thus, we mathematically define it as: 

\begin{equation}
\frac{\text{\# of correct pred.}}{\text{\# of correct} + {\text{incorrect}} + \text{missing pred.}}
\end{equation}

\textbf{Entity Frequency} captures how may entities are being mentioned in the focus areas, that are useful for predicting the communities. This is motivated by the fact that good focus areas should mention detailed topics for $\text{LLM}_{\text{task}}$ to focus on. For simplicity, our goal is to have at least three useful entities in the focus areas. Mathematically, let $g_e$ be number of entities mentioned more in one of the gold communities vs. another, and let $f_e$ be the number of entities mentioned in the focus areas and in $g_e$. Then, the reward is: $min(1.0 , f_e / 3)$.

\textbf{Focus Area Informativeness} scores the focus areas using a pre-trained model from ChatGPT data. Thus, to compute the reward: Let LR be the Regression model scoring info and f be the focus area. Then, the reward is: $s_f = LR(f)$

\textbf{Focus Area Length} aims to make focus areas longer in length, so that they are potentially more detailed. To compute it, Let $f_w$ be the number of words in the focus area, then the reward is: 0.5 if $f_w$ < 10, 1.0 if $f_w$ > 35, else $\frac{f_w - 10}{35 - 10} * (1.0 - 0.5) + 0.5$

\section{Curriculum Learning and Training Details}
\label{app:curriculum_learning}
\subsection{Training Details}
\label{appendix:training_details}
We provide training details in this section. 

We train our T5-base model using the public repository published by \citeauthor{akyurek2023rl4f} and \cite{ramamurthy2022reinforcement}. Our models are trained using a 12GB Titan XP GPU card, and intial supervised training takes 1 day. Subsequently, future RL training iterations also take one day. We make calls to the OpenAI ChatGPT API, using the models available publicly in November 2023, at the time these experiments were performed. For Llama 2, we run a local model, with 70B parameters, published by the Llama-cpp-python library\footnote{\url{https://github.com/abetlen/llama-cpp-python}}.

We used the development set to evaluate model performance, and choose the best hyper-parameters for our experiments.

As our prompt model, we train the T5-base model with a max prompt length of 650, for 120 epochs, a 0.00001 learning rate, and weight decay 0.01. For the RL stage, we fine-tune the T5-base model with all the same parameters, but a learning rate of 0.0001, entity coefficient of 0.1 and target KL of 3.

For downstream evaluation (news media profiling: news source political bias/factuality detection), our entire graph (train, test, and dev sets) has 2,969,854 edges, 81,326 nodes, 1,468 source nodes, and 35,099 user nodes.

\subsection{Curriculum Learning}
We use curriculum learning to learn our novel reward functions from Sec.~\ref{subsubsec:reward_functions}. We do curriculum learning, as averaging the different rewards into one reward score, and using that one score throughout the training process, makes learning each reward difficult. This is because the model cannot separate between the rewards, as it only gets one score, so it can't learn each reward function individually, and performance suffers.

However, there are benefits to using multiple rewards, as evidenced in the RL literature \cite{dann2023reinforcement}. Particularly, in our case, we want focus areas to be informative, capture relevant entities, be detailed, and be useful for community detection. Thus, we designed our reward functions to capture this, so when we optimize these rewards, the focus areas have these properties. This leads to focus areas being useful for community detection, and without these rewards, they wouldn't be.

Using curriculum learning, we learn each reward function, one at a time, to ensure the model can optimize each one. We introduce an additional reward function once model performance does not improve on the validation set for three training iterations. We first optimize for downstream performance (coverage), and then entity frequency. While these rewards lead to our model producing useful focus areas with relevant entities, they are relatively short and not detailed enough to separate users into accurate communities. Thus, once performance doesn’t increase on the validation set for three iterations, we add in the informativeness and finally length reward functions. As all the reward functions are added individually and used until performance stalls, they can be learned by the model, and they expand the initial focus areas to be more detailed and longer (thus also more useful). Once reward functions are used, they contribute equally to the final reward score (when compared to existing reward functions). However, as they are added sequentially, the model can still optimize them. We also use an additional reward, ROUGE score, which always contributes 25\% to the final reward. This reward scores the generated focus areas using the ROUGE metric and the gold data, to make sure the model continues to generate focus areas that are grammatically sound.

In this way, curriculum learning helps us optimize all of our reward functions, learning focus areas that are useful for community detection. We additionally performed an ablation study on the individual reward functions in Sec.~\ref{subsec:ablation_study}, which showed that while each reward function improves performance, learning them together through curriculum learning does the best.

\section{Gold Focus Area Generation}
\label{appendix:gold_focus_area_generation}
In this section, we discuss how we generate the gold focus areas to train $\text{LLM}_{\text{prompt}}$ in the initial supervised learning stage. To do this, we take advantage of the fact that we know the gold communities. We use $\text{LLM}_{\text{task}}$ to generate the gold focus areas, as we hope to initialize our $\text{LLM}_{\text{prompt}}$ model to the performance of $\text{LLM}_{\text{task}}$. Further, using $\text{LLM}_{\text{task}}$ to generate focus areas instead of humans allows us to quickly generate training data for a large amount of samples, which would otherwise be cost expensive.

\begin{figure*}[t!]
  \centering
  \includegraphics[scale=0.5]{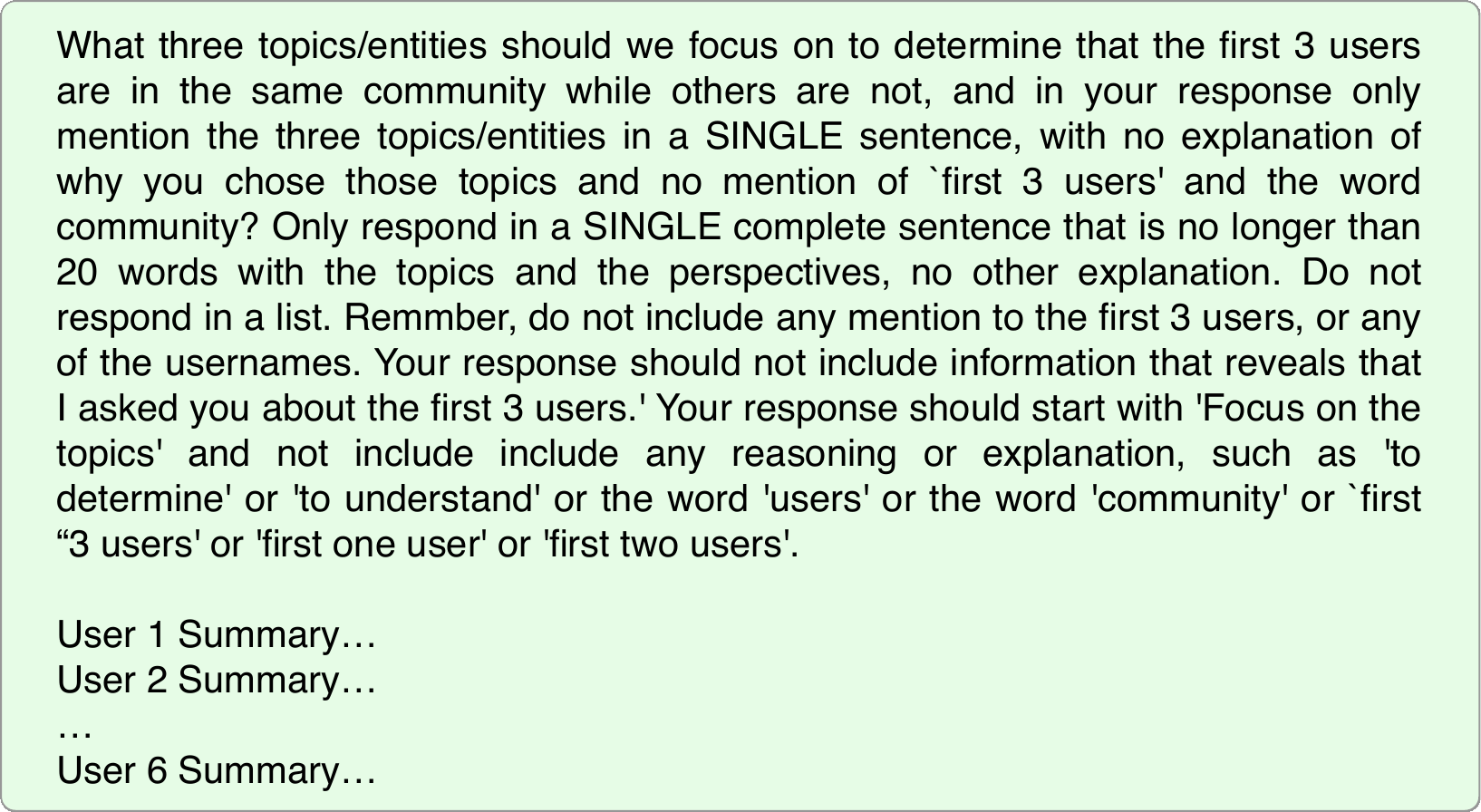}
\caption{\small An example of the prompt we used to generate gold focus areas. Given a set of six users, we ask $\text{LLM}_{\text{task}}$ what makes the first three users part of the same community. We also add additional instructions to the prompt to make sure that the LLM responds only with focus areas, not extra information such as user ordering.}
\label{fig:llm_task_generate_gold_focus_areas}
\vspace{-10pt}
\end{figure*}

Specifically, we prompt $\text{LLM}_{\text{task}}$ to separate the communities given the user summaries. For this, we provide the users to $\text{LLM}_{\text{task}}$ in sorted order (all users from first community first, all users from the second community second), asking it to provide the topics/entities to separate them. As $\text{LLM}_{\text{task}}$ often generates extra text that should not be part of focus areas and also often mentions the ordering of the users (which will not be valid at test time since the users will be randomly ordered), we additionally provide extra instructions in the prompt to try and avoid this. The exact question we ask is shown in Fig.~\ref{fig:llm_task_generate_gold_focus_areas}.

\subsection{Llama 2 Results}
\label{appendix:llamam2_results}
In this section, we provide results for our models when using Llama 2 \cite{touvron2023llama} as $\text{LLM}_{\text{task}}$, instead of ChatGPT as used in the main paper. All other settings are the same as when we used ChatGPT. Results are shown in Table~\ref{table:llama_results}, and show similar trends to using ChatGPT, showing our framework generalizes across different $\text{LLM}_{\text{task}}$ models. While the improvements of Llama 2 with focus areas are not as significant as ChatGPT, due to the fact that the Llama 2 model is not strong enough to take full advantage of focus areas, we still see significant improvements, showing the usefuleness of our framework.

\section{Discuss Cont: Human Analysis}
\label{app:human_analysis}
In this section, we continue our discussion from Sec.~\ref{subsec:human_analysis} and provide more details of our human analysis process. 

The goal of this step is to evaluate our focus areas, and determine if the focus areas generated by our framework are better than the ones produced by ChatGPT. While Sec.~\ref{sec:experiments} shows that this is the case across a variety of community detection and downstream tasks, in this section we have humans evaluate this.

To do this, we show three human annotators 50 samples (each human sees all 50). Each sample has one focus area from ChatGPT, and another generated by our best $\text{LLM}_{\text{prompt}}$ RL model for ChatGPT. For each sample, the human is asked to compare the focus areas, and then score them on a scale of 1-5, for grammatical correctness and usefulness. The usefulness rating identifies how useful the human believes the focus area will be to determine information communities. Ideally, a useful focus area should focus on divisive issues. The exact question we ask them is: Given two sentences (focus areas), score each on a scale of 1-5 (1 being lowest, 5 highest) for grammatical correctness and usefulness. The usefulness rating should capture how useful the focus area is to determine information communities. Ideally, a useful focus area should focus on divisive issues. The grammatical correctness rating should capture how grammatically correct the focus area is.

Results showed that while ChatGPT is more gramatically correct (4.95 vs 3.00), $\text{LLM}_{\text{prompt}}$ generates more useful focus areas (3.26 vs 3.07) across the 50 samples. This validates our experimental settings, where $\text{LLM}_{\text{prompt}}$'s focus areas lead to higher downstream performance, because they are more useful and focus the model on divisive issues to appropriately separate user communities.

The human annotators we used for this experiment were 20-30 year old male Ph.D. students in Computer Science and NLP, who are not authors of the paper or familiar with the study before the interaction process. One was Asian-American, one was Indian, and one was American. The students were provided fair working conditions and rewarded with research credit hours for their work in performing this annotation.

\section{Discussion: Case Study}
\label{sec:case_study}
In this section, we analyze our model by performing several case studies. We start by providing examples of high and low quality focus areas in Sec.~\ref{subsec:high_low_quality_focus_areas}, making it clearer what we want our focus areas to looks like. Then, in Sec.~\ref{subsec:chat_vs_llm_focus_areas}, we analyze the focus areas our trained model generates vs. ChatGPT, showing the benefit of our supervised and RL training procedure. Finally, in Sec.~\ref{subsec:focus_areas_help} we show detailed examples of how focus areas improve community detection performance, showing snippets of user summaries and how the communities formed are better once focus areas are used.

\begin{table*}[ht!]
\begin{center}
\begin{tabular}{|p{7.5cm}|p{7.5cm}|}
  \hline
  {\textbf{\small Low Quality Focus Areas}} & {\textbf{\small High Quality Focus Areas}} \\
 \hline
  \small Focus on Donald Trump. & \small Focus on Donald Trump's views on gun control. \\
  \small Focus on political opinions and perspectives. & \small Focus on Democrats. Republican lawmakers and their perspectives. Republican lawmakers and their treatment of immigrant. \\ 
  \small Focus on Fox News. & \small Focus on Sean Hannity's suitability as a diplomat. \\
 \hline
\end{tabular}
\caption{\small Examples of ``high quality'' and ``low quality'' focus areas, based on our definition. High quality focus areas tell the model what divisive issues/important topics and entities to focus on, so it can better detect the information community.}
\label{tab:focus_area_examples}
\end{center}
\end{table*}

\subsection{High and Low Quality Focus Areas}
\label{subsec:high_low_quality_focus_areas}
We aim for focus areas to tell the bigger LLM, $\text{LLM}_{\text{task}}$, exactly what topics to focus on. Ideally, focus areas shouldn't be about high level issues, but rather divisive topics. In Table~\ref{tab:focus_area_examples}, we provide examples of high and low quality focus areas. All of these were generated by the $\text{LLM}_{\text{task}}$ models presented in our framework. Note that the higher quality focus areas focus on issues, rather than just high level entities, which is what enables focus areas to lead to better community detection. Moreover, they mention relevant entities, are informative, and are detailed, due the fact that we trained with several relevant reward functions (see Sec.~\ref{subsubsec:reward_functions}). 

\subsection{ChatGPT vs \texorpdfstring{$\text{LLM}_{\text{task}}$}{LLM task} Focus Areas}
\label{subsec:chat_vs_llm_focus_areas}
Table.~\ref{tab:focus_areas_prompt_gpt} shows several examples of focus areas generated by our $\text{LLM}_{\text{prompt}}$ model and ChatGPT. From this, we can see that our $\text{LLM}_{\text{prompt}}$ model generates more useful focus areas, as they inform $\text{LLM}_{\text{task}}$ exactly of the topics and divisive issues to focus on to detect user information communities. This qualitatively shows the benefit of our Supervised + RL training procedure.

\subsection{Focus Areas Improving Community Detection}
\label{subsec:focus_areas_help}
\begin{figure*}[t!]
  \centering
  \includegraphics[scale=0.5]{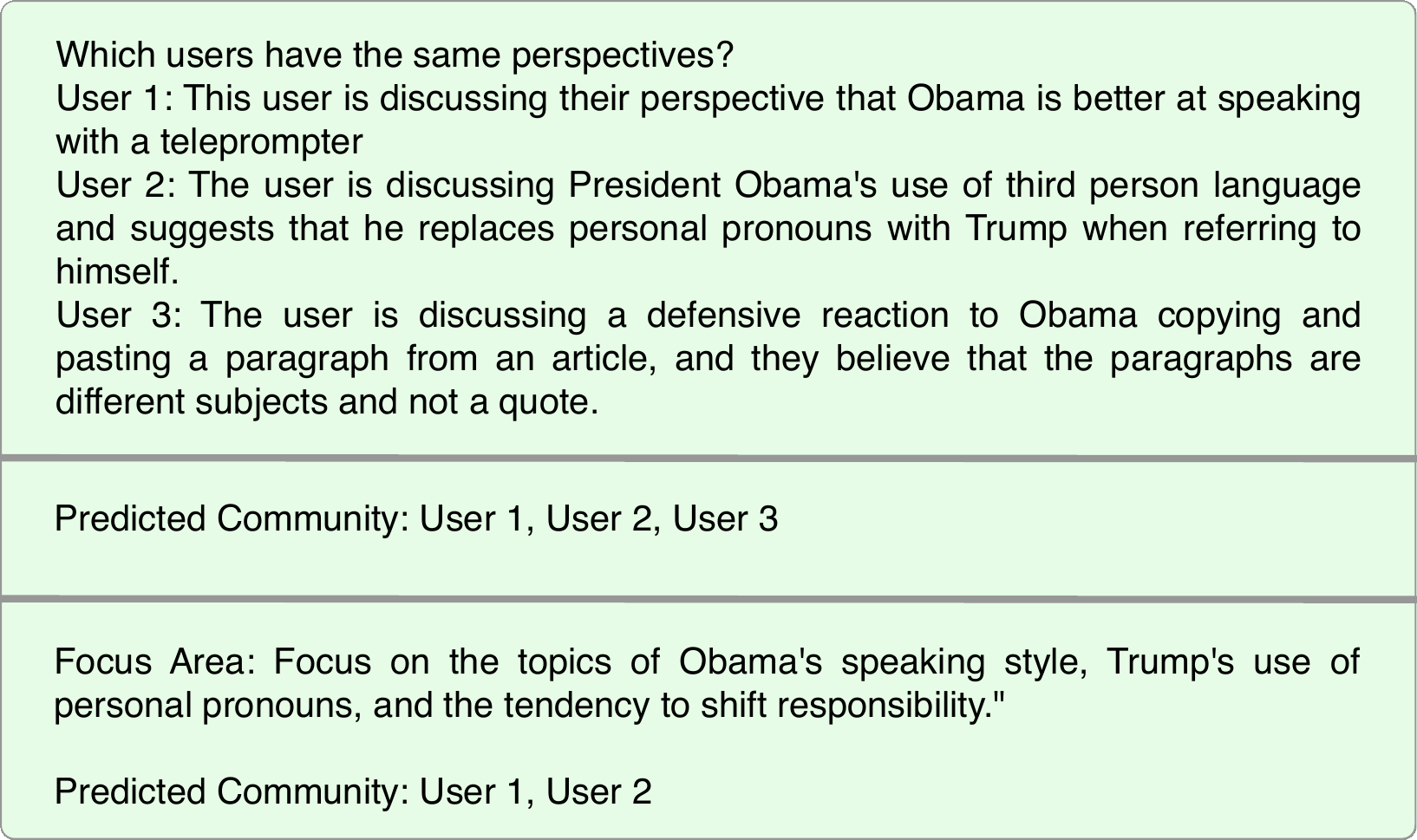}
\caption{\small Success case: An example of how focus areas can improve community detection. We show a few users and a snippet of their summaries, in sorted order for clarity. Without focus areas, the LLM predicts that all three users should be in the same community, as they all discuss President Obama's speech. However, when asked to focus on Obama's speaking style by the focus area, the LLM correctly identifies that Users 1 and 2 are similar as they criticize Obama's speech, while User 3 is defensive of his speech.}
\label{fig:improving_comms_1}
\vspace{-10pt}
\end{figure*}

\begin{figure*}[t!]
  \centering
  \includegraphics[scale=0.5]{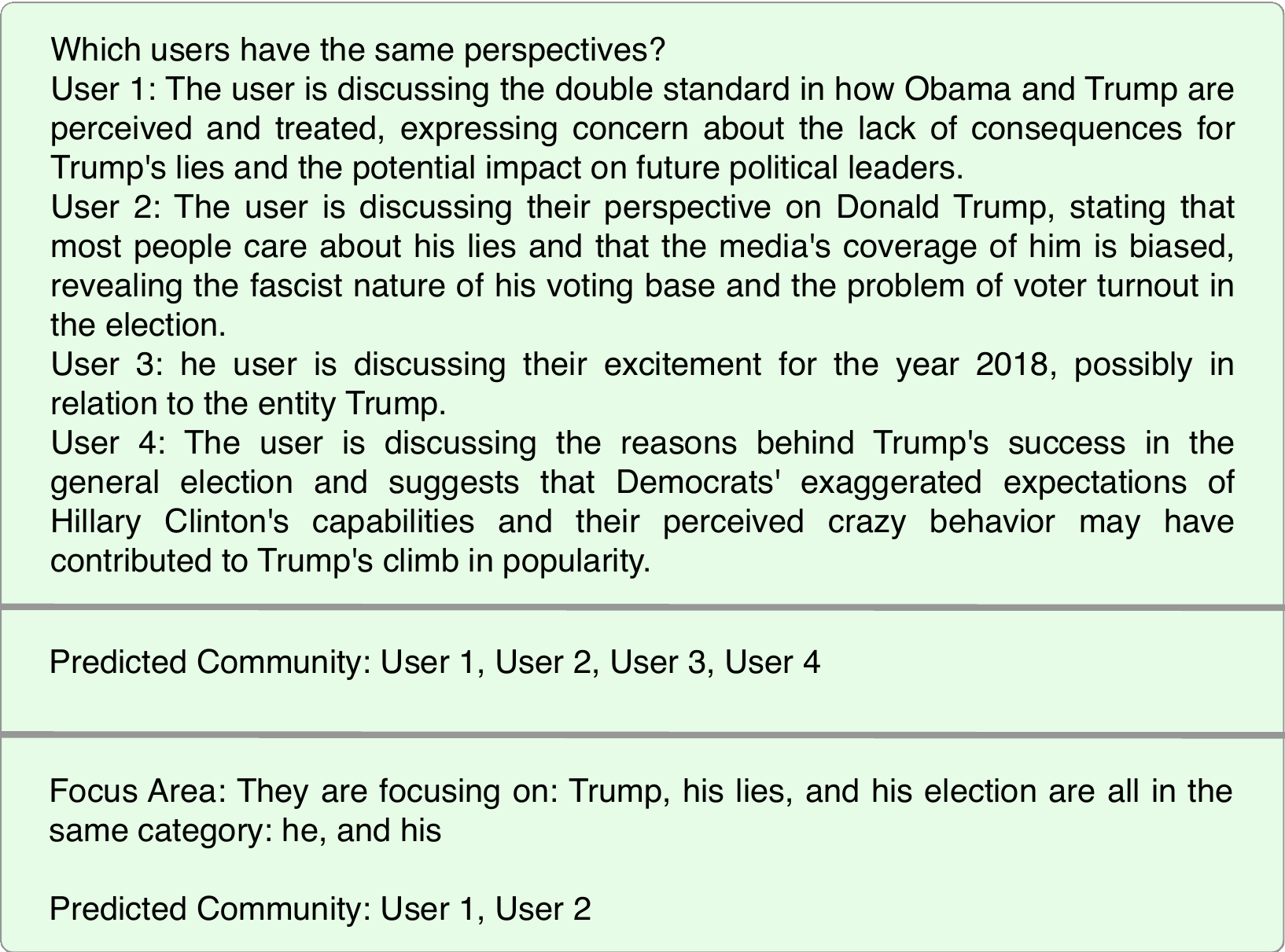}
\caption{\small Success case: An example of how focus areas can improve community detection. We show a few users and a snippet of their summaries, in sorted order for clarity. Without focus areas, the LLM can't correctly predict the community, as they all discuss President Trump. However, when asked to focus on President Trump's lies, it's clear that the first two users are against Trump, and the LLM can predict it correctly.}
\label{fig:improving_comms_2}
\vspace{-10pt}
\end{figure*}

\begin{figure*}
  \centering
  \includegraphics[scale=0.5]{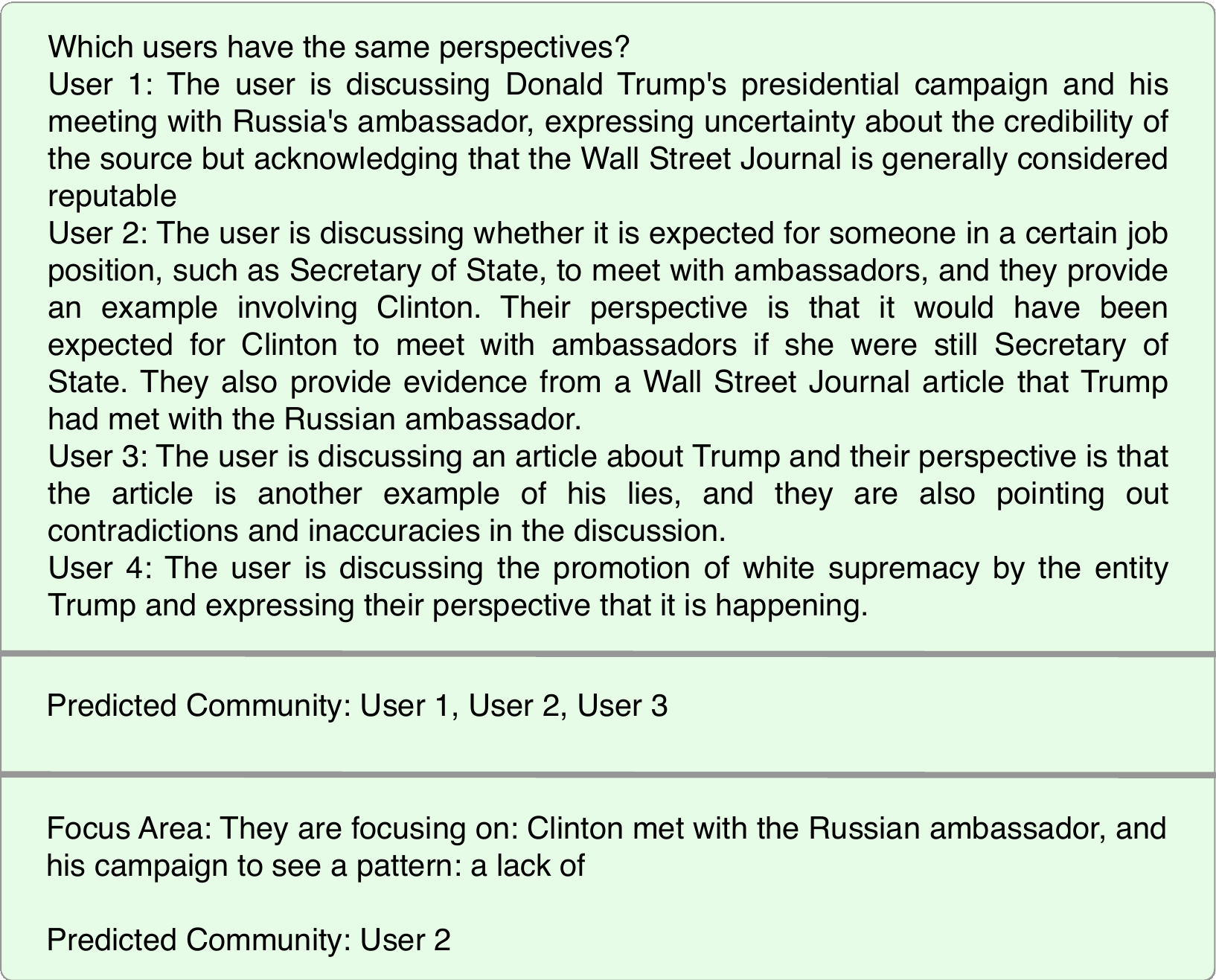}
\caption{\small Failure case: An example of how focus areas can hurt community detection. We show a few users and a snippet of their summaries, in sorted order for clarity. In this case, the focus area is too specific, leading to a one user community being formed, which is not very impactful.}
\label{fig:hurting_comms_1}
\vspace{-10pt}
\end{figure*}

Fig.~\ref{fig:improving_comms_1} and Fig.~\ref{fig:improving_comms_2} shows cases where focus areas can help improve community detection performance. On the contrary, Fig.~\ref{fig:hurting_comms_1} shows a case where focus areas can hurt community detection, if they are too specific (like in this case), or if they are too high-level/random (not shown).

\begin{table*}[ht!]
\begin{center}
\begin{tabular}{|p{7.5cm}|p{7.5cm}|}
  \hline
  {\textbf{\small ChatGPT Focus Area}} & {\small \textbf{$\text{LLM}_{\text{prompt}}$ Focus Area}} \\
 \hline
  \small Focus on the Governor of Virginia's campaign ad. & \small Focus on social entitlement and equality for conservatives in America. diverse demographic in the White House. \\ 
  \small Focus on clean coal. & \small Focus on death threats on Twitter. Free-Market Republicans. the entity Twitter and its \\ 
  \small Focus on the Republican party. & \small Focus on Donald Trump and his perspective on the deal with North Korea. Donald Trump \\ 
  \small Focus on political opinions and perspectives. & \small Focus on Democrats. Republican lawmakers and their perspectives. Republican lawmakers and their treatment of immigrant. \\ 
  \hline 
  \small Focus on the topic of America and its current state. & \small Focus on Michael Savage's perspective on tying social media accounts to people's \\ 
 \hline
\end{tabular}
\caption{\small Examples of focus areas generated by ChatGPT and our best RL + Curriculum Learning $\text{LLM}_{\text{prompt}}$ model. The first section shows cases where the human annotator from Sec.~\ref{subsec:human_analysis} believed $\text{LLM}_{\text{prompt}}$ was better, and the second section where they rated ChatGPT to be better.}
\label{tab:focus_areas_prompt_gpt}
\end{center}
\end{table*}

\section{Discussion: Learned Communities for News Source Factuality Detection}
\label{sec:discussion_graph_user_purity}
In this section, we analyze how well $\text{LLM}_{\text{task}}$ with focus areas allows us to learn communities that are relevant for the downstream task of news source factuality detection. To do this, we cluster (K-means, k=17) graph user embeddings before and after the $\text{LLM}_{\text{task}}$ communities are added into the graph (as discussed in Sec.~\ref{subsubsec:twitter_data}), and evaluate cluster purity.  

Specifically, we clustered the users in the test set graph before and after LLM-based communities are created (i.e. before and after the new user edges based on the $\text{LLM}_{\text{task}}$ communities are added to the graph), and evaluated the cluster purity. To compute purity, each cluster is assigned to the class which is most frequent in the cluster, and then the accuracy of this is measured. We assign labels to the users by propagating directly downwards from the source factuality labels (i.e. a user that follows 3 high factuality sources and 1 tweets 1 low factuality article has a label “high” factuality). We  cluster user graph embeddings, from the trained graph model, but do not do any training after the communities are created using the LLM.

The results show that user purity improves $\sim 3\%$, from 55.22 before to 58.66 after $\text{LLM}_{\text{task}}$ communities are added to the graph, showing that the communities formed by the LLM are meaningful, as users with similar factuality labels cluster closer together.

\section{Discussion: Impact of RL}
\label{app:rl_impact}
In this section, we further discuss the impact of the Reinforcement Learning (RL) stage on many of the results presented in the main paper, showing why this stage is crucical to both our community detection and downstream task performance.

Specifically, when compared to the Supervised Learning approach when using ChatGPT as the LLM, Reddit Political improves from 45.48\% to 47.85\%, a > 5\% relative performance improvement, and Reddit Economic improves from 44.00\% to 45.58\%, a > 3\% relative performance improvement. All of this improvements is on unseen data from future time periods/topics, compared to the training set. We hypothesize that additional RL rewards, such as improving the grammar of the focus areas, could also improve performance more.

RL is also critical to our Downstream task evaluation on Fake News Detection and Political Bias Detection. Here, we compared to SOTA models that outperform multiple baselines (SVM, GNN, Trained Text Classifier, etc.). Our results show the benefit of building communities, and without the RL stage of our approach, this improvement would not be possible. The model without RL would perform worse than existing baselines on this downstream task.

Finally, the RL stage also leads to better focus areas (due to rewards like Focus Area Entities), which is important for the real-world deployment of our approach. Thus, RL is a critical component of our approach. Among other benefits, it leads to performance improvements and better focus areas.

\end{document}